\documentclass[sigconf]{acmart}
\usepackage{algorithm}
\usepackage{algorithmic} 

\AtBeginDocument{%
  \providecommand\BibTeX{{%
    \normalfont B\kern-0.5em{\scshape i\kern-0.25em b}\kern-0.8em\TeX}}}

\copyrightyear{2021} 
\acmYear{2021} 
\setcopyright{acmcopyright}\acmConference[MM '21]{Proceedings of the 29th ACM International Conference on Multimedia}{October 20--24, 2021}{Virtual Event, China}
\acmBooktitle{Proceedings of the 29th ACM International Conference on Multimedia (MM '21), October 20--24, 2021, Virtual Event, China}
\acmPrice{15.00}
\acmDOI{10.1145/3474085.3475549}
\acmISBN{978-1-4503-8651-7/21/10}

\begin{document}

\title{Video Similarity and Alignment Learning on \\ Partial Video Copy Detection}

\author{Zhen Han,\quad Xiangteng He*,\quad Mingqian Tang,\quad Yiliang Lv}

\makeatletter
\def\authornotetext#1{
\if@ACM@anonymous\else
    \g@addto@macro\@authornotes{
    \stepcounter{footnote}\footnotetext{#1}}
\fi}
\makeatother
\authornotetext{Corresponding author.}

\affiliation{
 \institution{\textsuperscript{\rm}Alibaba Group, Hangzhou, China}
 \country{}
 }
\email{{hanzhen.hz, xiangteng.hxt, mingqian.tmq, yiliang.lyl}@alibaba-inc.com}

\def\authors{Zhen Han, Xiangteng He, Mingqian Tang and Yiliang Lv}

\renewcommand{\shortauthors}{Zhen Han, et al.}

\begin{abstract}
Existing video copy detection methods generally measure video similarity based on spatial similarities between key frames, neglecting the latent similarity in temporal dimension, so that the video similarity is biased towards spatial information.
There are methods modeling unified video similarity in an end-to-end way, but losing detailed partial alignment information, which causes the incapability of copy segments localization.
To address the above issues, we propose the Video Similarity and Alignment Learning (VSAL) approach, 
which jointly models spatial similarity, temporal similarity and partial alignment.
To mitigate the spatial similarity bias, we model the temporal similarity as the mask map predicted from frame-level spatial similarity, where each element indicates the probability of frame pair lying right on the partial alignments.
To further localize partial copies, the step map is learned from the spatial similarity where the elements indicate extending directions of the current partial alignments on the spatial-temporal similarity map. Obtained from the mask map, the start points extend out into partial optimal alignments following instructions of the step map.
With the similarity and alignment learning strategy, VSAL achieves the state-of-the-art $F_{1}$-score on VCDB core dataset. Furthermore, we construct a new benchmark of partial video copy detection and localization by adding new segment-level annotations for FIVR-200k dataset, where VSAL also achieves the best performance, verifying its effectiveness in more challenging situations. Our project is publicly available at \textcolor{blue}{\url{https://pvcd-vsal.github.io/vsal/}}.
\end{abstract}


\maketitle

\begin{figure}[!t]
  \centering
  \includegraphics[width=\linewidth]{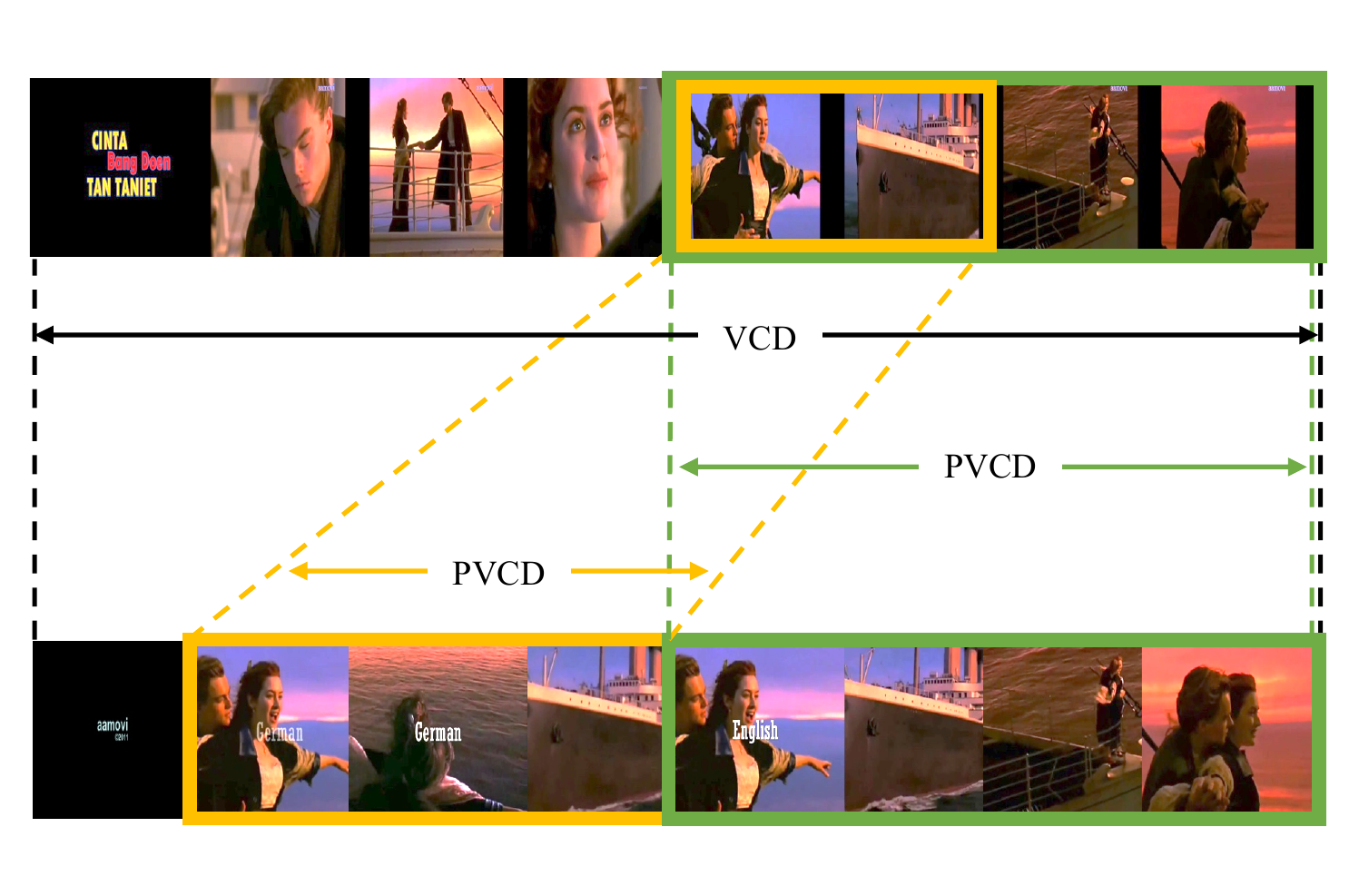}
  \caption{The illustration of Video-level Copy Detection (VCD) and Partial Video Copy Detection (PVCD). VCD aims to discover copy videos and PVCD focuses on localize both video-wise and segment-wise copies.}
  \Description{Pipeline of VSAL.}
  \label{PIPELINE}
\end{figure}

\section{Introduction}
In the past few decades, video applications such as video producing, re-producing and distributing have become convenient and low-cost. As a result, video copyright infringement has been extremely common on social medias and video-sharing platforms, making video copyright protection important and urgent. To solve these problems, \emph{Video-level Copy Detection (VCD)} aims to discover the copies from a large scale of video database. 
In practical application, it is necessary to know which part of the video is the copy. \emph{Partial Video Copy Detection (PVCD)} \cite{jiang_partial_2016} is such paradigm to not only discover the video-wise copies, but also localize the segment-wise copy video clips for a specific query video. The difference between VCD and PVCD is shown in Figure \ref{PIPELINE}.
Since some of those copies have been edited extensively and transformed significantly introducing huge spatial and temporal differences with the origins, making partial video copy detection a very challenging task. 
An effective PVCD system is required to have two capabilities: \emph{similarity measurement} and \emph{partial alignment localization}.

For similarity measurement, existing methods generally based on frame-level representation, such as local descriptor based representation \cite{zhang2016effective}, color correlation \cite{guzman2016towards}, DCT coefficients \cite{guzman2019partial} and CNN deep feature \cite{jiang_partial_2016, kordopatis2017near, baraldi2018lamv, shao2021temporal, self-supervised2021}. The above representations mainly focus on the spatial information and neglect the temporal information, which causes that the video similarity is easily biased towards spatial information of the key frames. 
To take the temporal information into consideration, a verification step called temporal alignment is often adopted, such as Dynamic Programming \cite{chou2015pattern, liu2017image}, Temporal Networks \cite{tan2009scalable, jiang2014vcdb, wang2017compact} and Temporal Hough Voting \cite{douze2010image, jiang_partial_2016}. 
Temporal alignment filters out the falsely matched frame pairs, and then the final video similarity is calculated by aggregating the frame-level similarities of the matched frame pairs.
However, it is still dominated by the spatial information encoded in separated representations, leaving the temporal similarity that hidden in the temporal order of frame-level similarities still underestimated. Facing varies spatial and temporal copy transformations, relying on any single dimension would lead to the lack of robustness. Therefore it is crucial to take both spatial and temporal similarities into consideration for video similarity measurement.

To explore the effect of spatial and temporal structure of the visual similarity, some end-to-end similarity aggregation methods, e.g. ViSiL \cite{kordopatis2019visil}, learn a single spatio-temporal similarity from the frame-to-frame similarity matrix, but the black-box models discard detailed sequence alignment information, which causes the incapability of partial copy alignment localization. This is also the common problem that the existing methods face to.

To effectively and simultaneously model similarity measurement and partial alignment localization, this paper proposes the Video Similarity and Alignment Learning (VSAL) approach. Its contributions can be summarized as follows: 
\begin{itemize}
\item
\textbf{Mask-based temporal similarity measurement}:
We propose a novel representation of the temporal similarity called mask map, which is learned from frame-to-frame spatial similarity map. Each element of mask map indicates the probability of each possible frame pair lying right on a partial alignment. In this way, information hidden in the temporal order of frame-level similarities is modeled jointly with the spatial similarity, generating spatio-temporal similarity to reduce the spatial bias of video similarity measurement. 

\item
\textbf{Step-based partial alignment}:
We further propose a prediction of the step map based on frame-to-frame spatial similarity map, which indicates partial alignments extending directions, and makes a lead to complete the alignments as long as the correct start points are provided. In this way, the detailed alignment information for copy segments is jointly modeled with spatio-temporal similarity to optimize the video similarity measurement comprehensively.
\end{itemize}

Besides, extensive experiments on VCDB and FIVR-200k-PVCD verify the effectiveness of the proposed VSAL approach, achieving the best performances. Especially,  we construct a new benchmark of partial video copy detection, called FIVR-200k-PVCD, by adding new segment-level annotations on FIVR-200k, which further evaluate the effectiveness of the proposed VSAL approach in more challenging situations.

\section{Related Work}

The relevant video copy detection tasks mainly include Near-Duplicate Video Retrieval (NDVR)\cite{wu2007practical, song2013effective, jiang2019svd}, Duplicate Scene Video Retrieval (DSVR)\cite{kordopatis-zilos_fivr_2019, kordopatis2019visil} and Partial Video Copy Detection (PVCD)\cite{jiang_partial_2016}. 

From the aspect of detection strategy, retrieval tasks NDVR and DSVR are the so-called Video-level Copy Detection (VCD) that require systems to rank copy videos ahead of the irrelevant ones. For efficient video similarity measurement. Most retrieval methods have a straightforward motivation aggregating the local frame-level features into clip-level or even video-level representations, such as global vectors\cite{wu2007practical, kordopatis2017near}, hash codes\cite{song2011multiple, jiang2019svd, self-supervised2021}, Bag-of-Words (BoW)\cite{cai2011million, kordopatis2017nearbow, liao2018ir}, and video similarity is measured by distances of aggregated representations. However, the aggregated representations are too coarse to cover abundant fine-grained information and can't be used to partial segment localization. Therefore in this paper, we adopt frame-level representation of SVRTN\cite{self-supervised2021} to implement frame encoding, keeping the fine-grained information for partial segment localization.

From the aspect of video copy type, copy videos in NDVR dataset like CC\_WEB\_VIDEO\cite{wu2007practical} and UQ\_VIDEO\cite{song2011multiple} are near-duplicate videos which are close to duplicate of each other but different in terms of photometric and editing variations, encoding parameters, file format, etc. Copies in DSVR and PVCD dataset like FIVR-200k\cite{kordopatis-zilos_fivr_2019} and VCDB\cite{jiang_partial_2016} are more complicated partial copies but DSVR only evaluates video-level detection and only provides the video pair annotations. Therefore in this paper, we construct a new FIVR-200k-PVCD benchmark by adding copy segment annotations to the DSVR subset of FIVR-200k\cite{kordopatis-zilos_fivr_2019} which contains more challenging spatial and temporal transformed copy segments.

\section{Video Similarity and Alignment Learning}
\begin{figure*}[h]
  \centering
  \includegraphics[width=\linewidth]{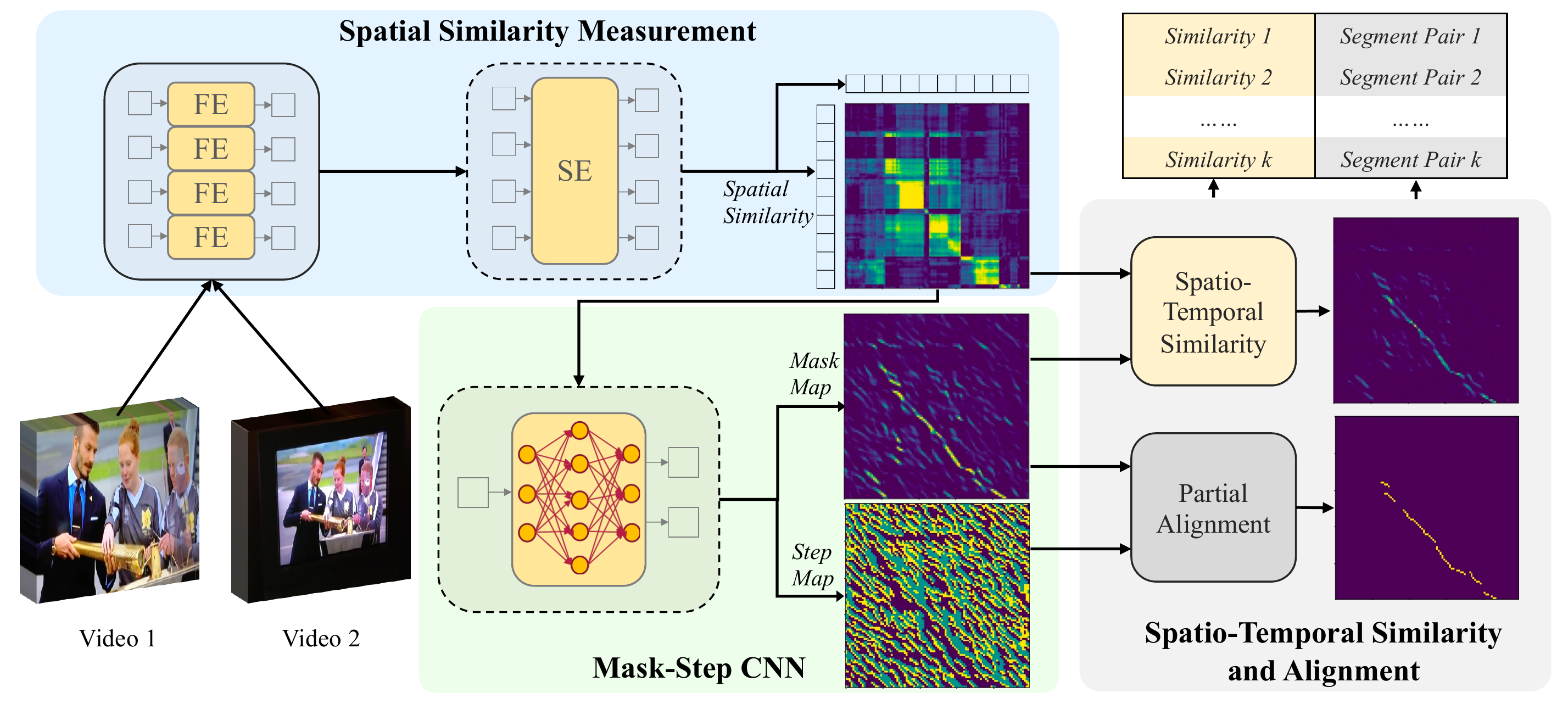}
  \caption{Overview of proposed VSAL approach. Spatial similarity measurement encodes input frames into frame-level representations and measure the spatial similarity. Learning from the spatial similarity, Mask-Step CNN predicts mask map and step map. MM together with spatial similarity and SM produce spatio-temporal similarity and partial alignment.
  Modules within the dashed box are jointly learning modules.}
  \Description{Two branch model.}
  \label{ARCH}
\end{figure*}
In this section, we formulate the video similarity into the combination of three components: spatial similarity, temporal similarity and partial alignment. Based on this similarity measurement, we propose the Video Similarity and Alignment Learning (VSAL) approach to jointly model these three components, as shown in Figure \ref{ARCH}.
It consists of three modules:
(1) Spatial similarity measurement is based on frame-level representations to generate the spatial similarity. 
(2) Mask-Step CNN aims to predict the mask map (MM) and step map (SM) simultaneously, which represent temporal similarity and an instruction map to complete partial alignments respectively. 
(3) Spatio-temporal similarity and alignment is based on spatial similarity, mask map and step map to calculate spatio-temporal similarities and localize copy segments. 

\subsection{Problem Formulation}
Given two input video sequences $u,v$ containing $M$, $N$ frames respectively, there are three components of the similarity between $u$ and $v$ which are spatial similarity ($S$), temporal similarity ($T$) and partial alignment ($P$). Therefore, the video similarity ($Sim$) can be formulated as:
\begin{equation}
Sim = \mathcal{F}(S, T, P)
\end{equation}
The formal description of $S$, $T$ and $P$ are presented below.

\subsubsection{Spatial Similarity Matrix}
\noindent

\begin{figure}[!t]
  \centering
  \includegraphics[width=\linewidth]{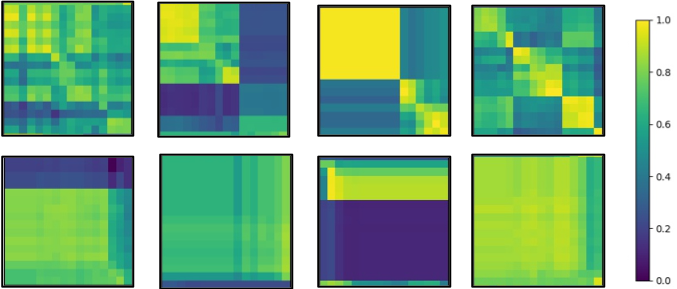}
  \caption{Examples of spatial similarity matrices. The upper row shows the spatial similarity matrices of well-matched video pairs and others are matrices of unmatched video pairs. All the matrices are trimmed to 16$\times$16.}
  \Description{VSAL pipeline.}
  \label{PATTERN}
\end{figure}

\noindent
Spatial similarity is calculated only based on individual frame content, denoted as a matrix $S=(s_{i,j})\in\mathbb{R}^{M\times N}$, where the elements are frame-level similarities between frame pairs. Some examples of $S$ are visualized in Figure \ref{PATTERN}.

\subsubsection{Partial Alignment}
\noindent

\noindent
As illustrated in Figure \ref{PATTERN}, spatial similairty matrices of well-matched videos often have one or more clear consecutive diagonal paths as patterns of sequence alignments where frame-level similarities are much greater than the off-path ones. The diagonal paths are composed of consecutive pairwise matched frames, generally following the temporal order, but partial alignments do not neccesarily follow temporal order all the time and could begin or end at any position in the spatial similairty matrices.We use a set $P_{k}$ to denote the $k$-th partial alignment between $u$ and $v$, where its elements are frame pairs marked by the frame ids $(i, j)\in\mathbb{N}_{+}^{2}$. In VASL, a step map is learnt from $S$ to model $P$ and the detialed method is presented in Section \ref{CNN}.

The effect of partial alignmen can be described from two perspectives:
(1) It filters out false matched frames, which are not the true copy frame pairs.
(2) The length of $P_k$ indicates the confidence of a detection proposal as a copy segment. Previous works generally drop the copy segment proposals which is shorter than a threshold, called hard weight (HW) method. On the contrary, we adopt a soft weight (SW) as confidence score weighted on the final video similarity. Since proposals with longer time span should have greater confidence, SW is formulated by an increasing function as follows:
\begin{equation}
  \alpha_{k}=\frac{1}{1+\gamma e^{-\|P_{k}\|}}
  \label{WEIGHT}
\end{equation}
where $\|P_{k}\|$ denotes the shorter length of two segments corresponding to $P_{k}$  and $\gamma$ is a temperature parameter. It is easy to notice that $\alpha_k$ increases as $P_{k}$ geting longer and it is approaching its upper bound 1 when $P_{k}$ is long enough.

\subsubsection{Temporal Similarity Matrix}
\noindent

\noindent
Unlike spatial similarity is calculated only based on individual frame content, temporal similarity mainly focuses on the strength of temporal alignment, also denoted as a matrix $T=(t_{i,j})\in\mathbb{R}^{M\times N}$. 
Some temporal arrangement information has been taken into account when extracting $P_{k}$. However, the temporal arrangement information is totally lost in the similarity calculation by simply collecting the frame-level similarities located in $P_k$. As a consequence, the final similarity is usually biased towards spatial information in frame-level representations. As the examples present in Figure \ref{PATTERN}, despite of the various strengths of spatial similarity, we can also see the different degrees of temporal alignment between well-matched videos and unmatched videos. Therefore, the temporal similarity needs to be estimated and can also be learned from spatial similarity matrix. A mask map is also learnt from $S$ to model $T$ and the detialed method is presented in Section \ref{CNN}.

\subsubsection{The Final Video Similarity}
\noindent

\noindent
As all three components are fully represented, we calculate the final video similarity by jointly considering $S$, $T$ and $P$. Similarity $Sim_{k}$ of the $k$-th partial alignment can be defined as:
\begin{equation}
  Sim_{k}=\frac{\alpha_{k}}{|P_{k}|}\sum_{i,j\in P_{k}}{s_{i, j}t_{i, j}}
  \label{SIMILARITY}
\end{equation} 
where $|\cdot|$ denotes the cardinality of a set. Based on this formulation, we construct a model to jointly model above three components.

\subsection{Spatial Similarity Measurement}
In spatial similarity measurement, we first extract frame-level representations from spatial encoders including a frame encoder (FE) and a sequence encoder (SE), and then calculate the spatial similarity based on these frame-level representations. 

Specifically, we uniformly sample frames from $u$ and $v$, and features are extracted by a single pre-trained image feature encoder individually forming two frame-level feature sequences $U\in\mathbb{R}^{M\times W}$, $V\in\mathbb{R}^{N\times W}$, where $W$ is the frame-level feature dimension. 
Then a transformer\cite{vaswani2017attention} encoder layer $f_{\theta}$ is deployed as the sequence encoder to enhance the frame-level representation via interaction between those individual spatial information.
Through sequence encoding, sequence length and vector dimension remain identical to the original $U$ and $V$.
Finally, spatial similarity $S$ are measured by matrix multiplication as follows:
\begin{equation}
  S = f_{\theta}(U)f_{\theta}(V)^{\mathsf{T}}
\end{equation}
Since U and V have been L2 normalized along dimension $W$, each element of $S$ is a cosine distance of a frame-level feature pair with corresponding timestamps.

\subsection{Mask-Step CNN}\label{CNN}
Based on the spatial similarity, we propose a Mask-Step CNN to simultaneously model the temporal similarity and partial alignment. 

\subsubsection{Architecture}
\noindent

\begin{table}
  \caption{Details of Mask-Step CNN architecture}
  \label{tab:commands}
  \begin{tabular}{cccc}
    \toprule
    Layer   & Kernel size/ & Output size & Activation\\
            &Padding&\\
    \midrule
    Conv-1 & $3\times3/1$ & $M\times N\times8$ & ReLU\\
    Conv-2 & $3\times3/1$ & $M\times N\times16$ & ReLU\\
    Conv-3 & $3\times3/1$ & $M\times N\times32$ & ReLU\\
    \midrule
    Mask & $3\times3/1$ & $M\times N\times2$ & Softmax\\
    \midrule
    Step & $2\times2/0$ & $(M-1)\times (N-1)\times3$ & Softmax\\
    \bottomrule
  \end{tabular}
  \label{CNN}
\end{table}
\noindent
As shown in Table \ref{CNN}, Mask-Step CNN is a two branch model with a mask branch and a step branch. The mask branch learns mask map (MM) as representation of temporal similarity, and step branch focuses on partial alignment by learning a step map (SM) which is a direction instruction map to find the potential partial alignment.

The backbone consists of 3 convolution layers followed by a nonlinear activation respectively. When spatial similarity $S$ passes through the backbone, spatial size of output feature maps $M$ and $N$ remain unchanged. 

For the mask branch, a 2-channel convolution layer is used to make a binary classification on each spatial position $(i,j)$, indicating the probability of corresponding position lying right on a partial alignment. The positions with higher probability are more likely the exact matched frame pairs. It is  noted that the mask map can also represent how well $s_{i,j}$ aligned along their temporal direction, so mask map is a representation of the temporal similarity $T$.

For the step branch, we also use a similar convolution layer for step predictor making a direction classification on each position, and the categories indicate directions to step next from current position to continue alignment path. Specifically, there are 3 options for each position stepping next which are corresponding to 3 category of classification: "stepping right-down", "stepping right" and "stepping down". Following the instructions of predicted step map with size of $(M-1)\times (N-1)$, we can easily walk through either $S$ or $T$ as long as an $M\times N$ map starting from an arbitrary position to build a path.

\subsubsection{Training Data and Label Generation}
\label{sec:trainingdata}
\noindent

\noindent
We train Mask-Step CNN and sequence encoder together in a self-supervised manner and both the training data and label are generated through data augmentation. We first collect a large amount of unlabeled videos from the Internet. Then, a data augmentation including temporal and spatial transformations is implemented to generate training data and label.

Four different temporal transformations including speed adjustment, freezing, deletion and do-nothing are optional and chosen randomly. The default frame rate of input sequences for training is 1 fps. For convenience of speed adjustment, we sample raw sequences in a higher frame rate of 2 fps, so the raw sequences are at 0.5$\times$ speed comparing to the default 1 fps sequence. For a target speed higher than 0.5$\times$, frames in the raw sequences are randomly removed to accomplish that speed. For example, if the target speed is 1$\times$, every 1 from 2 frames will be removed from raw sequence. For the target speed of 1.5$\times$, every 2 from 3 frames are removed. The rest target speeds can be done in the same manner. For freezing transformation, we randomly pick a frame and copy it several times. As for deletion, some consecutive frames are randomly picked and removed from original sequences. After temporal transformation, three types of spatial transformations similar to \cite{self-supervised2021} are considered: photometric variation, geometric transformation and other editing operations. Each training video is randomly transformed twice to a pair of sequences namely anchor and positive. 

\begin{figure}[!t]
  \centering
  \includegraphics[width=0.65\linewidth]{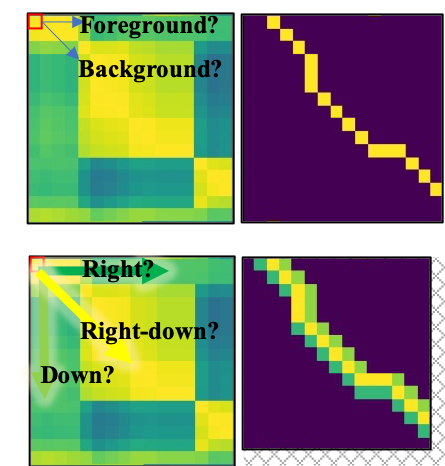}
  \caption{An example of mask label and step label. Mask label is above while step label is below.}
  \Description{Mask label and step label}
  \label{LABEL}
\end{figure}

After data augmentation, we generate training labels based on these transformations. During temporal transformation, all output frames are traced and each of them can be traced back to an original frame. For the $i$-th frame in anchor or positive sequence, $i^{\prime}$ denoting the id of corresponding original frame, which are noted as the clue to find the alignment path between anchor and positive. The alignment path can be denoted as $R=\{(i,j):i^{\prime}=j^{\prime}\}$, which means frames transformed from the same original frame are matched frames. Therefore the mask label can be obtained from $R$. The label of $(i,j)$ in mask map is 1 when $(i,j)\in R$ else 0. An example of mask label is shown in Figure \ref{LABEL} where yellow points indicate the labels in those positions are 1 and else 0.

For step branch, three categories (0, 1, 2) are predicted as three next step directions (right-down, right, down) respectively. However, the positions whose both above three directions' near neighbors are off the alignment path $R$ are considered not responsible for any direction prediction. Formally, only if $(i+1, j+1)$ or $(i+1, j)$ or $(i, j+1)$ in $R$, $(i,j)$ is responsible for predicting the next stepping direction. A set of the responsible $(i,j)$s and their target direction categories $l$ are defined following as $Q$:
\begin{equation}
Q=\left\{
             \begin{array}{l}
             \{(i,j,l):C_l(i,j)\in R\}, l=0 \\
             \{(i,j,l):C_l(i,j)\in R\ and\ C_0(i,j)\notin R \}, else.
             \end{array}
\right.
\end{equation}
where $C_l(\cdot)$ is the step function producing the next position according to category id $l$. Category 0, whose target direction is right-down, is the prior option. Categories of right and down are optional only if $C_0(i,j)\notin R$. According to $Q$, it is obvious that the target probability of each position $i,j$ and category $l$ can be defined as:
\begin{equation}
d_{i,j}^{l}=\left\{
             \begin{array}{l}
             0, (i,j,l)\notin Q \\
             0.5, l>0\ and\ (i,j,1)\in Q\ and\ (i,j,2)\in Q, \\
             1, else.
             \end{array}
\right.
\end{equation}
where superscripts $l$ denotes the category id.
In case of that both right and down are optional, we assign both categories a probability of 0.5. Therefore, $\sum_{l}{d_{i,j}^{l}}\equiv 1$. An example of step label is shown in Figure \ref{LABEL} where the category 0, 1 and 2 are represent by color yellow, dark green and light green respectively. The rest positions in the label map are not responsible positions and will not be involved in loss calculation.

\subsubsection{Multi-task Learning}
\noindent

\noindent

The overall loss is the combination of mask loss and step loss in a multi-task manner:
\begin{equation}
L=L_{m}+\lambda L_{s}
\end{equation}
where $L_m$ is mask loss, $L_s$ is step loss, and $\lambda$ is utilized to balance those two losses.

Mask loss is a Binary Cross-Entropy Loss that can be expressed as follows:
\begin{equation}
L_{m} = -\frac{1}{MN}\left(\sum_{(i,j)\notin R}{log(y_{i,j}^{0})} + \sum_{(i,j)\in R}{log(y_{i,j}^{1})}\right)\\
\end{equation}
where $y$ is the mask branch output, superscripts 0/1 denote the category id of background and foreground respectively, so the foreground probability $y^{1}$ is regarded as the element of MM representing temporal similarity $T$.

Step loss is also a Cross-Entropy Loss which is formulated as follows:
\begin{equation}
L_{s}= -\frac{1}{\sum_{(i,j,l)\in Q}{d_{i,j}^{l}}} \sum_{(i,j,l)\in Q}{log(z_{i,j}^{l})d_{i,j}^{l}}
\end{equation}
where $z$ is the predicted direction probabilities. Final decision are made by $max_{l}(z_{i,j}^l)$ where we obtain the so called SM.

\subsection{Spatio-Temporal Similarity and Alignment}
\begin{algorithm}[!t]
  \caption{Partial Alignment}
  \label{algopartalign}
  \begin{algorithmic}[1]
    \REQUIRE
    Spatial similarity $S=(s_{i,j})\in\mathbb{R}^{M\times N}$; Temporal similarity $T=(t_{i,j})\in\mathbb{R}^{M\times N}$; Step map $D=(d_{i,j})\in\mathbb{N}^{(M-1)\times (N-1)}$; Threshold to find start points $\tau$; Similarity threshold $\sigma$.
    \ENSURE Partial alignments $P$.
    \STATE $\Phi=\{(i,j): t_{i,j}>\tau\}$; $k=0$.
    \WHILE{$|\Phi|>0$}
    \STATE $k=k+1$.
    \STATE Set $P_k=\varnothing$; $g=0$.
    \STATE Select $(i,j)$ from $\Phi$ with smallest $i+j$ value.
    \WHILE{$i<M$ and $j<N$ and $g<3$}
    \STATE $st=s_{i,j}t_{i,j}$.
    \IF{$st<\sigma$}
    \STATE $g=g+1$.
    \ELSE
    \STATE Add $(i,j)$ to $P_k$.
    \ENDIF
    \STATE Remove $(i, j)$ and its 8 neighborhoods from $\Phi$.
    \STATE $(i,j)=C_{d_{i,j}}(i,j)$.
    \ENDWHILE
    \ENDWHILE
    \RETURN $P$.
  \end{algorithmic}
\end{algorithm}

From the spatial similarity measured by spatial representations, predicted mask map and step map, we identify partial alignments and then calculate a spatio-temporal similarity for every partial alignment. 

\subsubsection{Partial Alignment}
\noindent

\noindent
Partial alignment is implemented as shown in Algorithm \ref{algopartalign}.
We specify start point candidates as $(i,j)$s where $t_{i,j}>\tau$, and then starting from these points we keep stepping next following the instruction of SM until the map boundaries are reached or three consecutive $s_{i,j}t_{i,j}<\sigma$ occurred, where $\tau$ and $\sigma$ are predefined thresholds. Every time before stepping, the start point with the smallest index $i+j$ is chosen. To accelerate computation and avoid detecting an alignment repeatedly, start point candidates are removed if an alignment steps through one of their 8 neighborhoods. Moreover, only the matched frame pairs with spatio-temporal similarities larger than $\sigma$ are adopted to $P_k$. The alignment algorithm ends when no start point candidate remains. After detecting all the partial alignments, the start times and end times of corresponding copy segment pairs are obtained as the minimum and maximum timestamps of frame pairs in $P_{k}$.

\subsubsection{Spatio-Temporal Similarity}
\noindent

\noindent
At beginning, an element-wise multiplication between $S$ and $T$ is conducted to merge the spatial and temporal similarities, so we assign a spatio-temporal similarity to each frame pair. 
We collect corresponding spatio-temporal similarities of all frame pairs in $P_{k}$ and soft weight $\alpha_{k}$ are calculated as described in Equation \eqref{WEIGHT}. Finally, we calculate a spatio-temporal similarity $Sim_{k}$ for every partial alignment $P_k$, according to Equation \eqref{SIMILARITY}.

\section{Experiments}
\subsection{Implementation details}
For FE, we adopt the implementation of SVRTN$_f$\cite{self-supervised2021} from which we extract the float features and the feature dimension $W$ of frame-level representations including $U$,$V$,$f_{\theta}(U)$ and $f_{\theta}(V)$ are set to 512. For SE, we use a 8-head transformer encoder layer with 1024 hidden size and no positional embedding is added to the input features. SE are trained together with Mask-Step CNN. All features are L2-normalized along the feature dimension $W$ before and after SE.
 
For training data and label generation, we first sample raw frames from videos with a fixed rate of 2 fps. Four temporal transformations are randomly chosen with the same probability. For speed adjustment, the target speeds are randomly chosen from the range of 0.5$\times$ \textasciitilde 2.0$\times$ corresponding to frame rate of 2 fps \textasciitilde 0.5 fps. Before the transformations except speed adjustment, the frame rate is lowered to the default 1fps. In freezing and deleting transformations, at most 4 frames are repeated or removed. The output anchor and positive sequences are trimmed or padded into a fix length of 16 and they are required to have at least 4 frames matched.

For training process, we set $\lambda=1$ meaning that mask loss and step loss have the same weight. We adopt default SGD optimizer where momentum and weight decay are fixed to 0.9 and $10^{-5}$. The model is trained for 8 epoch with learning rate of $5\times 10^{-4}$ for the former 4 epochs and $5\times 10^{-5}$ for the last 4 epochs.

For the similarity calculation and partial alignment, the temperature parameter $\gamma$ in SW is set to 100. The start points threshold $\tau$ on mask map is fixed to 0.3 and $\sigma$ is set to 0.1 as threshold on spatio-temporal similarity.

\subsection{Datasets}
VSAL is trained in a self-supervised manner, so a large number of unlabelled raw videos are collected as our training data. For evaluation, a benchmark dataset called VCDB and a new labelled FIVR-200k are used.
\textbf{(1) Web Videos}: For self-supervised training, we collect videos about 3000 hours from video-sharing website. Videos are transformed temporally and spatially during training as explained in Section \ref{sec:trainingdata}.
\textbf{(2) VCDB}\cite{jiang_partial_2016}: VCDB dataset is constructed for partial video copy detection, which contains real-world videos of 28 different queries (Beckham 70 yard goal, dove evolution commercial, etc.) and 7 categories varying from speech to surveillance. The VCDB core dataset are published with 528 videos and annotations of over 9200 copy segment pairs both with start time and end time. Most segments are easy to detect and spatial and temporal transformations are relatively simple.
\textbf{(3) FIVR-200k-PVCD}: To evaluating performance on more complicated spatial and temporal situations, we add annotation of the segment pairs for DSVR subset of FIVR-200k\cite{kordopatis-zilos_fivr_2019} to construct the new partial video copy detection benchmark, called FIVR-200k-PVCD. Original FIVR-200k is a fine-grained instance video retrieval dataset consisting of 225,960 videos and 100 queries, including three retrieval tasks namely Duplicate Scene Video Retrieval (DSVR), Complementary Scene Video Retrieval (CSVR) and Incident Scene Video Retrieval (ISVR). Here we only focus on the annotations DSVR videos. Overall FIVR-200k-PVCD contains 10870 annotated copy segment pairs involving 5935 different video pairs. Many partial copy segments are more challenging with abundant temporal and spatial editing.

\subsection{Evaluation Metrics}
For VCDB evaluation, we follow the metrics in \cite{baraldi2018lamv, jiang_partial_2016}, namely segment-level precision (SP) and recall (SR) as well as the best $F_{1}$-score. As defined by VCDB benchmark, a segment pair both share frames to a ground-truth pair are considered as a correct detection.

Similar to VCDB, we also use SP, SR and the best $F_{1}$-score as evaluation metrics for FIVR-200k-PVCD. Additionally, we report them with different \textit{Intersection of Union} (IoU) constrains to better evaluate the precision of copy segments localization. As a prerequisite, a minimum IoU threshold is set for every proposed segment with groundtruth segment. A segment pair is considered as a correct detection only if both IoUs with a groundtruth pair are above the threshold.


\subsection{Comparisons with State-of-the-art Methods}
\begin{table}[!t]
  \caption{Comparison of segment-level performance between VSAL and other state-of-the-art methods on VCDB core dataset.}
  \label{tab:commands}
  \begin{tabular}{cccc}
    \toprule
    Methods                                     & SP        & SR        & $F_{1}$-score\\
    \midrule
    ATN\cite{jiang_partial_2016}                & 0.7050    & 0.5220    & 0.5956\\
    CNN\cite{jiang_partial_2016}                & -         & -         & 0.6503\\
    SNN\cite{jiang_partial_2016}                & -         & -         & 0.6317\\
    CNN+SNN\cite{jiang_partial_2016}            & -         & -         & 0.6454\\
    TH+CC+ORB\cite{guzman2016towards}           & 0.5052    & 0.9294    & 0.6546\\
    LAMV\cite{baraldi2018lamv}                  & -         & -         & 0.6870\\
    CNN+SC\cite{wang2017compact} (1fps)         & -         & -         & 0.6995\\
    CNN+SC\cite{wang2017compact} (all frames)   & -         & -         & 0.7038\\
    BTA\cite{zhang2016effective}                & 0.7600    & 0.7500    & 0.7549\\
    Q-Learning\cite{guzman2019partial}          &0.8829 & 0.7355 & 0.8025\\
    FPVCD\cite{knn2021}                           &-          &-          & 0.8613\\
    \textbf{VSAL}  &\textbf{0.8971}&\textbf{0.8462}&\textbf{0.8709}\\
    \bottomrule
  \end{tabular}
  \label{VCDB}
\end{table}
We compare the proposed VSAL approach with more than 10 state-of-the-art methods in VCDB core dataset benchmark and SP, SR and $F_{1}$-score are reported in Table \ref{VCDB}. 
These methods are briefly introduced below:
\textbf{ATN, CNN, SNN and CNN+SNN\cite{jiang_partial_2016}}: are reported by VCDB benchmark. ATN uses local descriptor based frame representations. CNN and SNN use deep features trained in different strategies. All the methods are implemented with spatial geometric verification and the Temporal Network as temporal alignment.
\textbf{TH+CC+ORB\cite{guzman2016towards}} uses two global representations TH and CC together with a local feature of ORB. The alignment is done with proposed multilevel matching.
\textbf{LAMV\cite{baraldi2018lamv}} is proposed to compare and align videos using temporal match kernels which finds temporal alignment in the Fourier domain.
\textbf{CNN+SC\cite{wang2017compact}} encodes key frames separately and then compress and aggregate features into a compact representation. Temporal Network is used for video segment matching.
\textbf{BTA\cite{zhang2016effective}} is proposed to search boundaries for partial copy segments with Binary Temporal Alignment method.
\textbf{Q-Learning\cite{guzman2019partial}} is a learning based method for copy video and segment decision which is adapted from a reinforcement learning technique.
\textbf{FPVCD\cite{knn2021}} is proposed to search with global feature and localize segments with modified temporal network.

As we can see, VSAL achieves the best performance on SP, SR and $F_{1}$-score comparing to others. Above all, VSAL is the only approach that takes the temporal similarity into similarity measurement which models the temporal order of frame-level similarities. Most methods introduce temporal information through alignment, such as Hough Voting and Temporal Network, making temporal information absence from similarity measurement. Other methods such as Q-Learning model pixel variances as temporal information, but such representations only encode the variances between two frames next to each other ignoring the long-range temporal relations. Moreover, VSAL utilizes a unified learning strategy that models spatial similarity, temporal similarity and partial alignment jointly via multitask learning, achieving better performance than other learning based methods such as Q-Learning, which only conducts learning on the decision stage. For partial alignment, step map learning from vast number of unlabelled data is more robust to deal with tricky temporal transformations than artificial rule-based methods used by BTA, CNN+SC and FPVCD.

\subsection{Ablation Study}
\begin{table}
  \caption{Ablation studies on VCDB core dataset.}
  \label{tab:commands}
  \begin{tabular}{l|ccc}
    \hline
    Methods     & SP & SR & $F_{1}$-score\\
    \hline
    HV (baseline)&0.8513&0.6912&0.7629\\
    HV+SE&0.8607&0.6936&0.7682\\
    HV+SE+SW&0.7686&0.7887&0.7785\\
    SM+SE+SW&0.8447&0.8047&0.8242\\
    SM+SE+SW+MM&\textbf{0.8971}&\textbf{0.8462}&\textbf{0.8709}\\
    \hline
  \end{tabular}
  \label{ABLATION}
\end{table}

To fully understand each module of VSAL, experiments are performed with different settings. We first construct a baseline model based on the spatial similarity calculated from frame encoder features and the alignment method of Hough Voting (HV)\cite{jiang_partial_2016,douze_compact_2010} with default hard weight (HW). Then four modules in VSAL including sequence encoder (SE), soft weight (SW), step map (SM) and mask map (MM) are added into the baseline model one after another. Performances are compared to verify the effectiveness of each module.

\begin{figure}[!t]
  \centering
  \includegraphics[width=\linewidth]{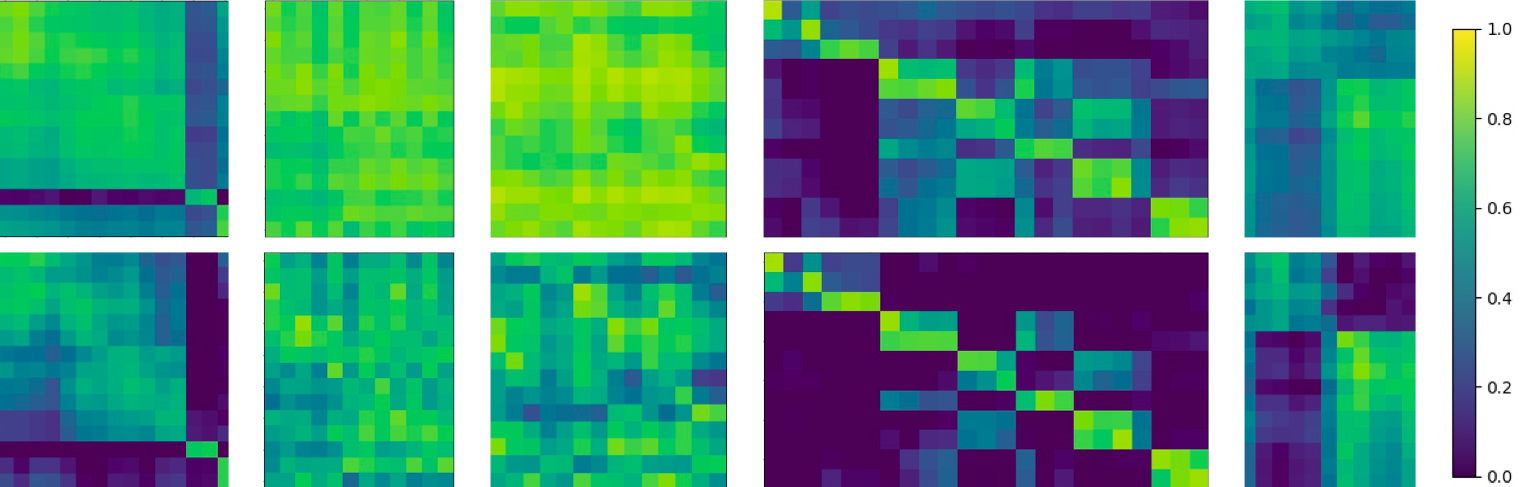}
  \caption{Comparison of spatial similarity matrices with or without sequence encoder. The spatial similarity matrices without sequence encoder are shown on the top while at the bottom row shows them with sequence encoder.}
  \Description{}
  \label{TRANSFORMER}
\end{figure}

\begin{figure}[!t]
  \centering
  \includegraphics[width=\linewidth]{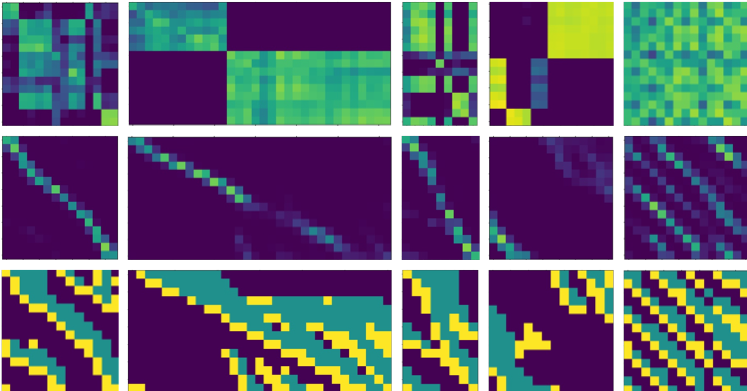}
  \caption{Visualization of Spatial Similarity and corresponding predicted MM and SM. The first row shows the spatial similarity and the two rows below shows the visualization of MM and SM respectively. In the last row, yellow, green and dark purple mean the different directions in SM: right-down, down and right respectively.}
  \Description{Spatial Similarity and corresponding predicted MM and SM.}
  \label{MMSM}
\end{figure}

\begin{table}[!t]
  \caption{Parameter sensitivity of parameters $\gamma$, $\tau$ and $\sigma$. $F_{1}$-scores are reported.}
  \label{tab:commands}
  \begin{tabular}{ccccc}
    \hline
    Parameters   & 0.1   & 0.2   & 0.3   & 0.4\\
    \hline
    $\tau$    &\textbf{0.8736} &0.8723 &0.8709 &0.8699\\
    $\sigma$  &\textbf{0.8709} &0.8649 &0.8189 &-\\
    \hline
    \hline
    Parameters   &10     &100    &500    &1000\\
    \hline
    $\gamma$ &0.8695 &\textbf{0.8709} &0.8664 &0.8650\\
    \hline
  \end{tabular}
  \label{para}
\end{table}

\begin{table*}[!t]
  \caption{Performance comparison on FIVR-200k-PVCD.}
  \resizebox{\textwidth}{15mm}
  {
  \begin{tabular}{l|ccc|ccc|ccc|ccc}
    \hline
                &       &IoU>0  &           &       &IoU>0.3&               &       &IoU>0.5&               &       &IoU>0.7&           \\     
    Methods     & SP    & SR    & $F_{1}$-score & SP&SR     & $F_{1}$-score & SP    &SR     & $F_{1}$-score & SP    &SR     & $F_{1}$-score\\
    \hline
HV(baseline) &0.4350 &0.5911 &0.5012 &0.6069 &0.3491 &0.4433 &0.5501 &0.3142 &0.4000 &0.4778 &0.2708 &0.3457 \\
HV+SE&0.4579 &0.5936 &0.5170 &0.5827 &0.3794 &0.4596 &0.5281 &0.3439 &0.4165 &0.5164 &0.2755 &0.3593 \\
HV+SE+SW&0.5730 &0.5255 &0.5483 &0.5075 &0.4563 &0.4805 &0.4541 &0.4128 &0.4325 &0.3952 &0.3553 &0.3742 \\
SM+SE+SW&0.8300 &0.5916 &0.6908 &0.8151 &0.5525 &0.6586 &0.7580 &0.5014 &0.6036 &0.6485 &0.4091 &0.5017 \\
SM+SE+SW+MM&\textbf{0.8575} &\textbf{0.6883} &\textbf{0.7636} &\textbf{0.8212} &\textbf{0.6556} &\textbf{0.7291} &\textbf{0.7738} &\textbf{0.5434} &\textbf{0.6384} &\textbf{0.7076} &\textbf{0.4281} &\textbf{0.5335} \\
    \hline
  \end{tabular}
  }
  \label{tab:FIVR}
\end{table*}

\subsubsection{Effectiveness of Sequence Encoder}
\noindent

\noindent
We first add the sequence encoder after the frame encoder and compare its performance with baseline setting. As shown in the first two rows of Table \ref{ABLATION}, we can observe that SE introduces a slight improvement of about 0.5\% $F_{1}$-score.

To better explain the improvement, visualized examples of spatial similarities calculated from FE and SE are presented in Figure \ref{TRANSFORMER}. As can be seen from the visualization, alignment paths become more clear by adding SE after FE. As a result, the jointly trained sequence encoder learns to enlarge the difference between frames as well as suppress the similar representations. It is especially helpful for finding partial alignments of motionless shots (as the first 3 columns in the figure) because minor differences between similar frames are easier to capture.

\subsubsection{Effectiveness of Soft Weight}
\noindent

\noindent
To demonstrate the effectiveness of soft weight (SW), we compare it with the default choice hard weight (HW). In HW, detected segments with length less than 3 are dropped to avoid false positives. In SW, in contrast, similarity is weighted by a confidence score which is calculated according to Equation \eqref{WEIGHT}. 
As shown in the 2nd and 3rd rows in Table \ref{ABLATION}, dealing with short segments in a soft way performs better in the best $F_{1}$-score than simply dropping results shorter than a certain length.

\subsubsection{Effectiveness of Step Map}
\noindent

\noindent
For the alignment stage, we replace HV with proposed learning based Step Map (SM) to evaluate its effectiveness. 
As shown in the 3rd and 4th rows, both SR and SP increase mainly because SM learning from vast number of unlabelled data is more robust facing critical temporal transformations. The improvement on VCDB is 4.5\% $F_{1}$-score. The effectiveness of step map will be further evaluated on FIVR-200k-PVCD in section \ref{FIVR} which contains more challenging situations. 
For SM, a visualization is presented in Figure \ref{MMSM}. 
As we can see, SM makes a decision of stepping direction for every single position except the last row and column. If a correct start points are selected, following the direction instructions we always get the correct alignment.

\subsubsection{Effectiveness of Temporal Similarity}
\noindent

\noindent
To demonstrate the improvement made by adding temporal similarity, we further utilize the model with mask map (MM). Final similarities are measured based on the spatio-temporal similarity that combines spatial and temporal similarity.
Performances are shown in the 4th and 5th rows of Table \ref{ABLATION}. By adding MM, a great improvement (4.6\% $F_{1}$-score) is achieved which proves that temporal similarity plays an important role in video copy detection and proposed MM is a good representation of the temporal similarity. 
As can be seen from the visualization of MM in Figure \ref{MMSM}, MM clearly shown the distribution of alignments and temporal similarities in some positions are very low despite of considerable high spatial similarities, e.g. MM at the second column and fourth column. It proves that temporal similarity represents hidden information which spatial similarity ignores.

\subsection{Parameter Sensitivity}
We implement an experiment to exploit the sensitivities of parameters $\gamma$, $\tau$ and $\sigma$. Performances are reported in Table \ref{para}. When $\tau$ drops from 0.3 to 0.1, $F_{1}$-score increases a little. However, the number of start points chosen for partial alignments grows over 90.5\%. To balance the efficiency and effectiveness, $\tau$ is set to 0.3. We set $\sigma$ to 0.1 and $\gamma$ to 100 for the best performance.

\subsection{Experiment on the New Benchmark FIVR-200k-PVCD}\label{FIVR}
We further evaluate the performance of VSAL on FIVR-200k-PVCD, which contains harder cases. As shown in Table \ref{tab:FIVR}, baseline model only gets about 50\% $F_{1}$-score on FIVR-200k-PVCD comparing to 76\% on VCDB, showing the difficulty of the new dataset.

Qualitatively, all the performance trends are the same as that on VCDB, but the quantitative improvements are much more obvious when replacing alignment method from HV to SM. As shown in Table \ref{tab:FIVR}, under all IoU thresholds, SM alignment achieves higher performance of over 10\% $F_{1}$-score. Comparing to HV alignment, SM has two advantages: flexible path selection and seeing further when making current decision. For every step, there are three optional directions making it robust dealing with critical temporal transformations. Moreover, as a benefit from CNN's large receptive fields, direction decisions are made referring to a large scale of neighbours to avoid short-sighted choices.

\section{Conclusion}
In this paper, we propose the VSAL method to jointly learn the spatial similarity, temporal similarity and partial alignment from unlabelled videos. Its novelty can be described as follows: (1) A novel representation of the temporal similarity is learned from the frame-level spatial similarity. The temporal information hidden in the order of spatial are well measured as so-called mask map, which can be combined with spatial similarity to generate a spatio-temporal similarity to balance the similarity measurement from both spatial and temporal aspects. (2) We model the partial alignment as a direction instruction map namely step map, which is also learned from the frame-level spatial similarity. The alignment information is jointly learned with spatio-temporal similarity which ensures the high performance and the capability of finding partial alignments. Experiments on VCDB and a more challenging new benchmark based on FIVR-200k, verify the effective of VSAL approach which achieves the best performance on both dataset.

The future works will be based on two aspect. Extreme temporal transformations like a recombined video of fine-divided shots will be taken into consideration. The divide-and-conquer strategy such as dividing long videos into pieces then merging the separate similarities and alignments together will be explored.

\begin{acks}
This work is supported by Alibaba Group through Alibaba Research Intern Program.
\end{acks}

\bibliographystyle{ACM-Reference-Format}
\bibliography{references}

\end{document}